\documentclass[10pt,twocolumn,letterpaper]{article}

\usepackage{iccv}
\usepackage{times}
\usepackage{epsfig}
\usepackage{graphicx}
\usepackage{amsmath}
\usepackage{amssymb}
\usepackage{caption}
\usepackage{pifont}
\newcommand{\xmark}{\ding{55}}%
\usepackage{booktabs} 
\usepackage{subfigure}

\usepackage[pagebackref=true,breaklinks=true,letterpaper=true,colorlinks,bookmarks=false]{hyperref}

\usepackage{diagbox}

\newcommand{\namefull}{Frequency Domain Image Translation}
\newcommand{\name}{FDIT}
\def\*#1{\mathbf{#1}}

\def\iccvPaperID{5504}

\iccvfinalcopy

\begin{document}

\title{Frequency Domain Image Translation: \\
More Photo-realistic, Better Identity-preserving}


\newcommand\distanceAuthor{0.5}
\newcommand\distanceInstitue{1.0}

\author{
Mu Cai$^1$\hspace{\distanceAuthor cm}Hong Zhang$^2$\hspace{\distanceAuthor cm}Huijuan Huang$^3$\hspace{\distanceAuthor cm}Qichuan Geng$^{4}$\hspace{\distanceAuthor cm}Yixuan Li$^1$\hspace{\distanceAuthor cm}Gao Huang$^{5}$
\vspace{0.2cm} \\ 
$^1$University of Wisconsin-Madison\hspace{\distanceInstitue cm} $^2$SenseTime Group Ltd. \hspace{\distanceInstitue cm} $^3$Kwai Inc. \\  $^4$Beihang University \hspace{\distanceInstitue cm} $^5$ Tsinghua University 
\\
{\tt\small \{mucai,sharonli\}@cs.wisc.edu\hspace{\distanceAuthor cm} fykalviny@gmail.com\hspace{\distanceAuthor cm} huanghuijuan@kuaishou.com} \\
\vspace{-0.02in}
{\tt\small zhaokefirst@buaa.edu.cn\hspace{\distanceAuthor cm} gaohuang@tsinghua.edu.cn}
}


\maketitle

\begin{abstract}

Image-to-image translation has been revolutionized with GAN-based methods. However, existing methods lack the ability to preserve the identity of the source domain. As a result, synthesized images can often over-adapt to the reference domain, losing important structural characteristics and suffering from suboptimal visual quality. To solve these challenges, we propose a novel frequency domain image translation (FDIT) framework, exploiting frequency information for enhancing the image generation process. Our key idea is to decompose the image into low-frequency and high-frequency components, where the high-frequency feature captures object structure akin to the identity. Our training objective facilitates the preservation of frequency information in both pixel space and Fourier spectral space. We broadly evaluate FDIT across five large-scale datasets and multiple tasks including image translation and GAN inversion. Extensive experiments and ablations show that FDIT effectively preserves the identity of the source image, and produces photo-realistic images. FDIT establishes \textbf{state-of-the-art} performance, reducing the average FID score by 5.6\% compared to the previous best method.

 \end{abstract}
\section{Introduction}

\begin{figure}[t]
	\begin{center}
	  \includegraphics[width=0.5\textwidth]{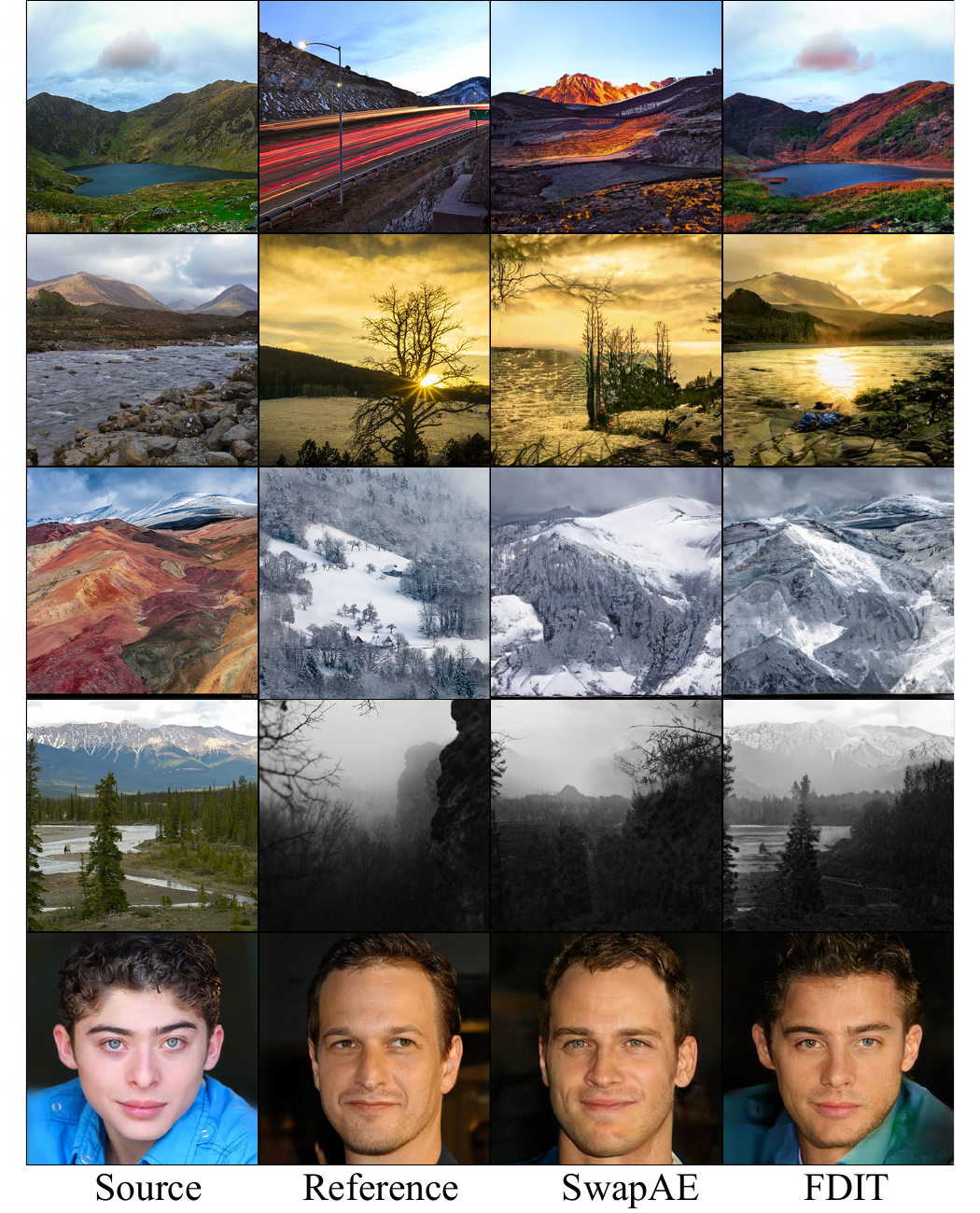}
	\end{center}
	\vspace{-4ex}
	   \caption{\small Image translation results of the Flicker mountains dataset. From left column to right: we show the source images, reference images, the generated images using Swapping Autoencoder~\cite{park2020swapping} and FDIT (ours), respectively. SwapAE over-adapt to the reference image. \name{} better preserves the composition and identity with respect to the source image.}
	   \vspace{-2ex}
	\label{fig:church_trans}
 \end{figure}

Image-to-image translation~\cite{zhu2017toward, chen2020domain, bhattacharjee2020dunit, xiong2020fine, wang2020attentive} has attracted great research attention in computer vision, which is tasked to synthesize new images based on the source and reference images (see Figure~\ref{fig:church_trans}). This task has been revolutionized since the introduction of GAN-based methods~\cite{8100115,CycleGAN2017}. In particular, a plethora of literature attempts to decompose the image representation into a content space and a style space~\cite{choi2020stargan, park2020swapping, DRIT, huang2018munit}. To translate a source image, its content representation is combined with a different style representation from the reference domain. 


Despite exciting progress, existing solutions suffer from two notable challenges. First, there is no explicit mechanism that allows preserving the identity, and as a result, the synthesized image can over-adapt to the reference domain and lose the original identity characteristics. This can be observed in Figure~\ref{fig:church_trans}, where Swapping Autoencoder~\cite{park2020swapping} generates images with identity and structure closer to the reference rather than the source image. For example, in the second row, the tree is absent from the source image yet occurs in the translation result.
Second, the generation process may lose important fine-grained details, leading to suboptimal visual quality. This can be prohibitive for generating photo-realistic high-resolution images. The challenges above raise the following important question: \emph{how can we enable photo-realistic image translation while better preserving the identity?}





\begin{figure}[bpt]
	\begin{center}
		\includegraphics[width=0.97\linewidth]{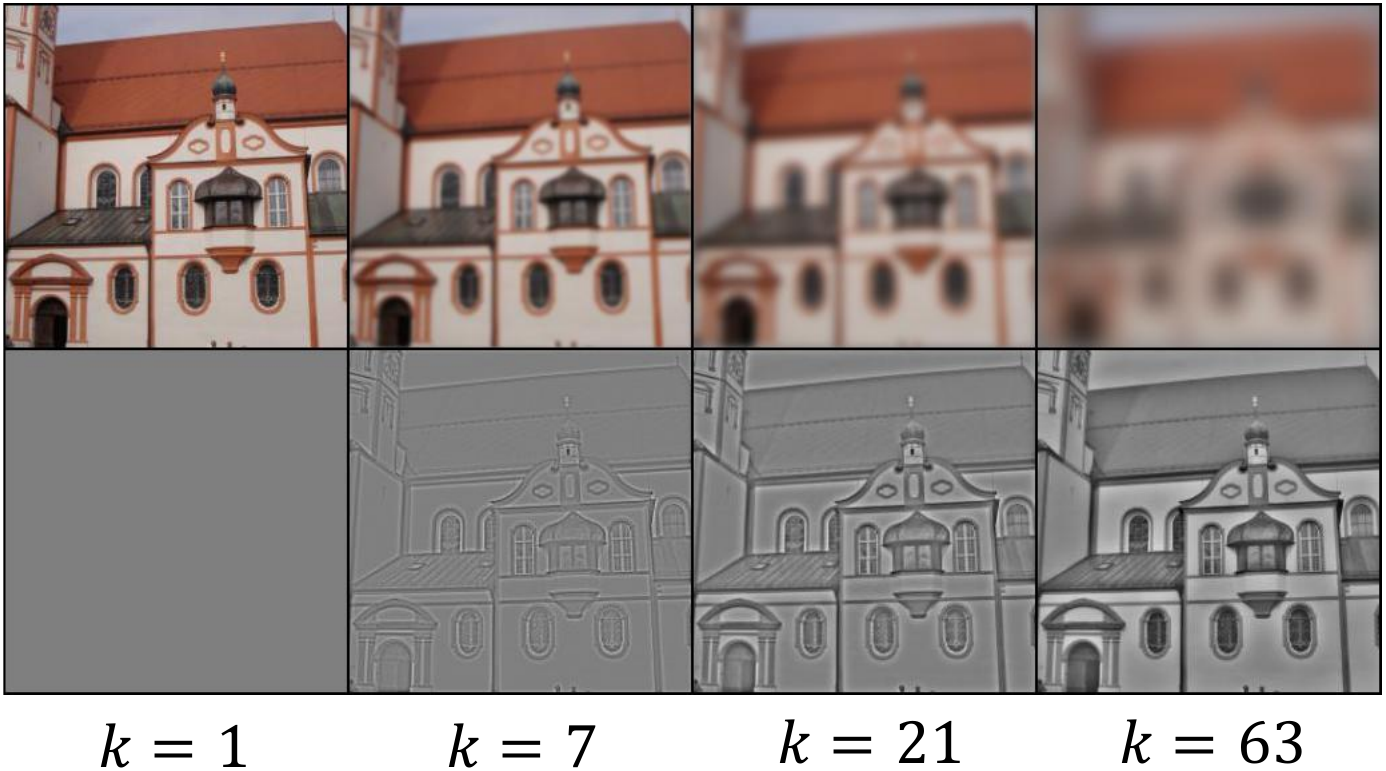}
	\end{center}
	\vspace{-2ex}
		\caption{\small Visualization of the effect of decomposing the original image into grayscale high frequency (\emph{bottom}) and low frequency (\emph{top}) components. Gaussian kernel is employed as the low-frequency filter with different kernel sizes $k$. 
		}
	   \vspace{-2ex}
	\label{fig:blur_compare}
 \end{figure}



Motivated by this, we propose a novel framework--\emph{\namefull{}}~(\textbf{\name{}})--exploiting frequency information for enhancing the image generation process. Our key idea is to decompose the image into low- and high-frequency components, and regulate the frequency consistency during image translation. Our framework is inspired by and grounded in signal processing~\cite{deng1993adaptive, brigham1988fast, heideman1984gauss}. Intuitively, the low-frequency component captures information such as color and illumination; whereas the high-frequency component corresponds to sharp edges and important details of objects. For example, Figure~\ref{fig:blur_compare} shows the resulting images via adopting the {Gaussian blur} to decompose the original image into low- \vs high-frequency counterparts (top \vs bottom). The building identity is distinguishable based on the high-frequency components.

Formally, \name{} introduces novel frequency-based training objectives, which facilitates the preservation of frequency information during training. The frequency information can be reflected in the visual space as identity characteristics and important fine details. Formally, we impose restrictions in both \emph{pixel space} as well as the \emph{Fourier spectral space}. In the pixel space, we transform each image into its high-frequency and low-frequency components by applying the Gaussian kernel (\ie, low-frequency filter). A loss term regulates the high-frequency components to be similar between the source image and the generated image. Furthermore, \name{} directly regulates the consistency in the frequency domain by applying Fast Fourier Transformation (FFT) to each image. This additionally ensures that the original and translated images share a similar high-frequency spectrum. 

Extensive experiments demonstrate that \name{} is highly effective, establishing \textbf{state-of-the-art} performance on image translation tasks. Below we summarize our
{key results and contributions}:

\begin{itemize}
    \item We propose a novel frequency-based image translation framework, FDIT,  which substantially improves the identity-preserving generation, while enhancing the image hybrids realism. \name{} outperforms competitive baselines by a large margin, {across all datasets considered}. Compared to the vanilla Swapping Autoencoder~(SwapAE)~\cite{park2020swapping}, \name{} decreases the FID score by \textbf{5.6\%}.

    \item  We conduct extensive ablations and user study to evaluate the  (1) identity-preserving capability and (2) image quality, where \name{} constantly surpasses previous methods. For example, user study shows an average preference of \textbf{75.40}\% and \textbf{64.39}\% for FDIT over Swap AE in the above two aspects.  We also conduct the ablation study to understand the efficacy of different loss terms and frequency supervision modules.

    \item      
    We broadly evaluate our approach across five large-scale datasets (including two newly collected ones). Quantitative and qualitative evaluations on  image translation and GAN-inversion tasks demonstrate the superiority of our method\footnote{Code and dataset are available at: \href{https://github.com/mu-cai/frequency-domain-image-translation}{https://github.com/mu-cai/frequency-domain-image-translation}}.
    
\end{itemize}



\begin{figure*}[bpt]
	\begin{center}
		\includegraphics[width=0.97\linewidth]{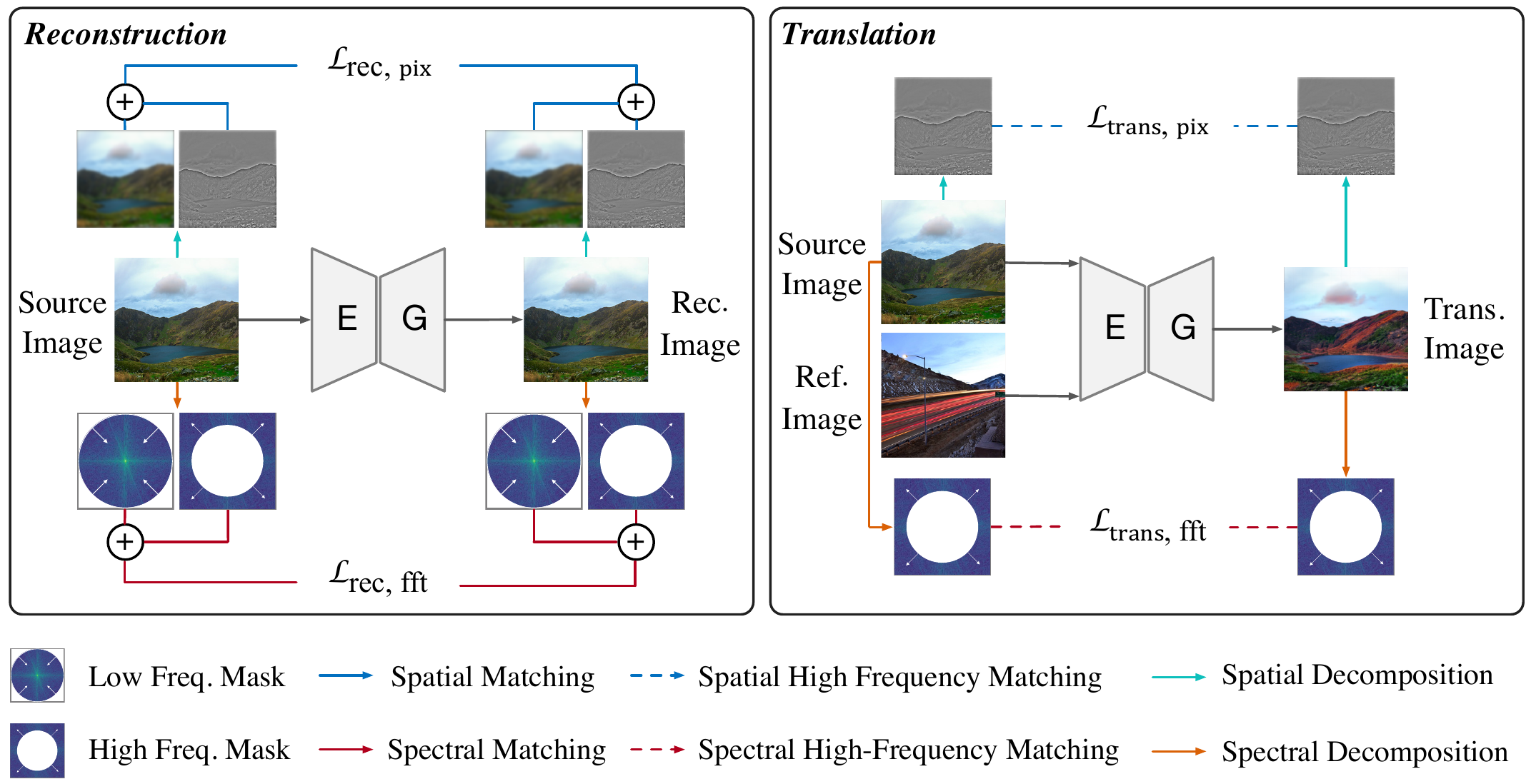}
	\end{center}
	\vspace{-4ex}
	   \caption{\small Overview of the proposed \emph{frequency domain image translation} (\textbf{FDIT}) framework. The key idea is to decompose the image into low-frequency and high-frequency components, and regulate the frequency consistency during image reconstruction (\emph{left}) and image translation (\emph{right}). High frequency information captures the sharp edges and important details of objects, where is effectively matched by FDIT training objectives.}
	   \vspace{-2ex}
	\label{fig:overall}
 \end{figure*}
 
 

\section{Background: Image-to-image Translation}
\label{sec:background}
Image-to-image translation aims at directly generating
the synthesized image given a source image and an accompanying
reference image.
Existing algorithms commonly employ an encoder-decoder–like neural network architecture. 
We denote the encoder $E(\mathbf{x})$, the generator $G(\mathbf{z})$, and the image space $\mathcal{X} = \mathbb{R}^{H \times W \times 3}$ (RGB color channels).  

Given an image $\mathbf{x} \in \mathcal{X}$, the encoder $E$ maps it to a latent representation $\mathbf{z} \in \mathcal{Z}$. 
Previous approaches rely on the assumption that the latent code can be composed into two components $\mathbf{z}=\left(\mathbf{z}_{c}, \mathbf{z}_{s}\right)$, where $\mathbf{z}_c$ and $\mathbf{z}_s$ correspond to the \textbf{c}ontent and \textbf{s}tyle information respectively. A reconstruction loss minimizes the $L_1$ norm between the original input $\mathbf{x}$ and $G(E(\mathbf{x}))$. 

To perform image translation, the generator takes the content code $\mathbf{z}_c^{\text{source}}$ from the source image, together with the style code $\mathbf{z}_s^{\text{ref}}$ from the reference image. The translated image is given by $G(\mathbf{z}_c^{\text{source}}, \mathbf{z}_s^{\text{ref}})$. However, existing methods can be limited by its feature disentanglement ability, where $\mathbf{z}_c^{\text{source}}$ may not capture the identity of source image. As a result, such identity-related characteristics can be undesirably \emph{lost in translation} (see Figure~\ref{fig:face_trans}), which motivates our work. 


\section{Frequency Domain Image Translation}

Our novel frequency-based image translation framework is illustrated in Figure~\ref{fig:overall}. In what follows, we first provide an overview and then describe the training objective. Our training objective facilitates the preservation of frequency information during the image translation process. Specifically, we impose restrictions in both \emph{pixel space} (Section~\ref{sec:pixel-loss}) as well as the \emph{Fourier spectral space} (Section~\ref{sec:fourier-loss}).


\subsection{Pixel Space Loss} 
\label{sec:pixel-loss}
\paragraph{High- and low-frequency images.} We transform each input $\*x$ into two images $\*x_{L}\in \mathcal{X}$ and $\*x_{H} \in \mathcal{X}$, which correspond to the low-frequency and high-frequency images respectively. Note that both $\*x_{L}$ and $\*x_{H}$ are in the same spatial dimension as $\*x$. Specifically, we employ the Gaussian kernel, which filters the high frequency feature and keeps the low frequency information:
\begin{equation}
	\begin{aligned}
		k_{\sigma}[i, j] &= \frac{1}{2 \pi \sigma^{2}} e^{-\frac{1}{2}\left(\frac{i^{2}+j^{2}}{\sigma^{2}}\right)},
	\end{aligned}
\end{equation}
where $[i, j]$ denotes the spatial location within the image, and $\sigma^{2}$ denotes the variance of the Gaussian function. Following~\cite{heideman1984gauss}, the variance is increased proportionally with the Gaussian kernel size . 
Using convolution of the Gaussian kernel on input $\*x$, we obtain the low frequency (\emph{blurred}) image $\mathbf{x}_L$:
\begin{equation}
	\begin{aligned}
		\mathbf{x}_L [i, j] &= \sum_{m} \sum_{n} k[m, n] \cdot \mathbf{x}[i+m, j+n].
	\end{aligned}
\end{equation}
where $m, n$ denotes the index of an 2D Gaussian kernel, \textit{i.e.},   $m, n \in [-\frac{k - 1}{2}, \frac{k - 1}{2}]$.




To obtain $\*x_{H}$,
we first convert color images into grayscale, and then subtract the low frequency information:
\begin{equation}
	\begin{aligned}
		\mathbf{x}_H &= { \texttt{rgb2gray} }(\mathbf{x})- (\texttt{rgb2gray}(\mathbf{x}))_{{L}},
	\end{aligned}
\end{equation}
where the $\texttt{rgb2gray}$ function converts the color image to the grayscale. This removes the color and illumination information that is unrelated to the identity and structure. The resulting high frequency image $\mathbf{x}_{{H}}$ contains the sharp edges, \ie \emph{sketch} of the original image. 

\paragraph{Reconstruction loss in the pixel space.} We now employ the following {reconstruction loss} term, which enforces the similarity between the input and generator's output, for both low-frequency and high-frequency components:
\begin{equation}
	\begin{aligned}
		\mathcal{L}_{\mathrm{rec, pix}}(E, G) 
		& = \mathbb{E}_{\mathbf{x} \sim \mathcal{X}}\Big[\|\mathbf{x}_{{L}}-(G(E(\mathbf{x})))_{{L}}\|_{1}\> \> \\
		& + \|\mathbf{x}_{{H}}-(G(E(\mathbf{x})))_{{H}}\|_{1}\Big].
	\end{aligned}
\label{eqn:adv_loss}
\end{equation}
\paragraph{Translation matching loss in the pixel space.} In addition to reconstruction loss, we also employ the {translation matching loss}:
\begin{small}
\begin{equation}
	\begin{aligned}
		\mathcal{L}_{\mathrm{trans, pix}}(E, G)=\mathbb{E}_{\mathbf{x} \sim \mathcal{X}}\left[\|\mathbf{x}^{\text{source}}_{{H}}-\left(G(\mathbf{z}_{c}^{\text{source}}, \mathbf{z}_{s}^{\text{ref}})\right)_{{H}}\|_{1}\right],
	\end{aligned}
	\label{eq:tranlation-loss}
\end{equation}

\end{small}
where $\mathbf{z}_{c}^{\text{source}}$ and $\mathbf{z}_{s}^{\text{ref}}$ are the content code of the source image and the style code of the reference image, respectively.
Intuitively, the translated images should adhere to the identity of the original image. We achieve this by regulating the high frequency components, and enforce the generated image to have the same high frequency images as the original source image. 

\subsection{Fourier Frequency Space Loss} 
\label{sec:fourier-loss}
\paragraph{Transformation from pixel space to the Fourier spectral space.} In addition to the pixel-space constraints, we introduce loss terms that directly operate in the Fourier domain space. In particular, we use Fast Fourier Transformation (FFT) and map $\mathbf{x}$ from the pixel space to the Fourier spectral space. We apply the Discrete Fourier Transform $\mathcal{F}$ on a real $2 \mathrm{D}$ image $I$ of size $H \times W$:
\begin{equation}
	\begin{aligned}
		\mathcal{F}(I)(a,b)=\frac{1}{HW} \sum\limits_{h=0}^{H-1} \sum\limits_{w=0}^{W-1} e^{-2 \pi i \cdot \frac{h a}{H}} e^{-2 \pi i \cdot \frac{w b}{W}} \cdot I(h, w), 
	\end{aligned}
\end{equation}
for $a=0, \ldots, H-1, b=0, \ldots, W-1$.

For the ease of post processing, we then transform  $\mathcal{F}$  from the complex number domain to the real number domain. Additionally, we take the logarithm to stabilize the training: 

\begin{equation}
	\begin{aligned}
		\mathcal{F}^R(I)(a, b)
		& = \mathrm{log}(1 + \sqrt{[\mathrm{Re}\mathcal{F}(I)(a,b)]^2} \\
		& + \sqrt{[\mathrm{Im}\mathcal{F}(I)(a,b)]^2} + \epsilon),
	\end{aligned}
\label{eqn:fft_log}
\end{equation}
where $\epsilon = 1\times 10^{-8}$ is a term added for numerical stability;  $\mathrm{Re}$ and $\mathrm{Im}$ denote the real part and the imaginary part of $\mathcal{F}(I)(a,b)$ respectively. Each point in the Fourier spectrum would utilize information from all pixels according to the discrete spatial frequency, which would represent the frequency features in the global level. 

\paragraph{Reconstruction loss in the Fourier space} We then regulate the {reconstruction loss} in the frequency spectrum:
\begin{small}
\begin{equation}
	\begin{aligned}
		\mathcal{L}_{\mathrm{rec, fft}}(E, G)=\mathbb{E}_{\mathbf{x} \sim \mathcal{X}}\left[\|\mathcal{F}^R(\mathbf{x})-\mathcal{F}^R(G(E(\mathbf{x})))\|_{1}\right].
	\end{aligned}
\label{equ:fft_recon}
\end{equation}
\end{small}
\paragraph{Translation matching loss in the Fourier space.} In a similar spirit as Equation~\ref{eq:tranlation-loss}, we devise a translation matching loss in the Fourier frequency domain:
\begin{small}
\begin{equation}
	\begin{aligned}
		\mathcal{L}_{\mathrm{trans, fft}}(E, G) &= \mathbb{E}_{\mathbf{x} \sim \mathcal{X}}\left[\| \mathcal{F}^R_{{H}} (\mathbf{x}^{\text{source}})  - \mathcal{F}^R_{{H}}(G\left(\mathbf{z}_{c}^{\text{source}}, \mathbf{z}_{s}^{\text{ref}}\right))\|_{1}\right],
	\end{aligned}
\end{equation}
\end{small}
where $\mathcal{F}^R_{{H}} (\mathbf{x}) =  \mathcal{F}^R({ \texttt{rgb2gray} }(\mathbf{x})) \cdot M_{{H}} $.  $M_{{H}}$ is the frequency mask, for which we provided detailed explanation below. The loss constrains the high frequency components of the generated images for better identity preserving.

\paragraph{Frequency mask.}

As illustrated in Figure~\ref{fig:overall}, the low-frequency mask is a circle with radius $r$, whereas the high-frequency mask is the complement region. The frequency masks $M_{{H}}$ and $M_{{L}}$ can be estimated empirically from the distribution of $\mathcal{F}^R$ on the entire training dataset. We choose the radius to be 21 for images with resolution 256$\times$256. The energy within the low-frequency mask accounts for 97.8\% of the total energy in the spectrum. 

\subsection{Overall Loss}
Considering all the aforementioned losses, the overall loss is formalized as:
\begin{equation}
	\begin{aligned}
		\mathcal{L}_{\text {FDIT }} &= \mathcal{L}_{\mathrm{org}} + \lambda_1 \mathcal{L}_{\mathrm{rec, pix}} + \lambda_2 \mathcal{L}_{\mathrm{trans, pix}} \\ & + \lambda_3\mathcal{L}_{\mathrm{rec, fft}} +  \lambda_4\mathcal{L}_{\mathrm{trans, fft}},
	\end{aligned}
\end{equation}
where $\mathcal{L}_{\mathrm{org}}$ is the orginal loss function of \textit{any} image translation model. For simplicity, we use $\lambda_1=\lambda_2=\lambda_3=\lambda_4=1$ in this paper. 

\paragraph{Gaussian kernel \vs FFT.} Gaussian kernel and FFT are complementary for preserving the frequency information. 
On one hand, the Gaussian kernel extracts the frequency information via the convolution, therefore representing the frequency features in a \emph{local} manner. On the other hand, Fast Fourier Transformation utilizes the information from all pixels to obtain the FFT value for each spatial frequency, characterizing the frequency distribution \emph{globally}.  Gaussian kernel and FFT  are therefore complementary in preserving the frequency information. We show ablation study on this in Section~\ref{sec:ablation}, where both are effective in enhancing the identity-preserving capability for image translation tasks.

\begin{figure}[h]
	\begin{center}
		\includegraphics[width=0.9\linewidth]{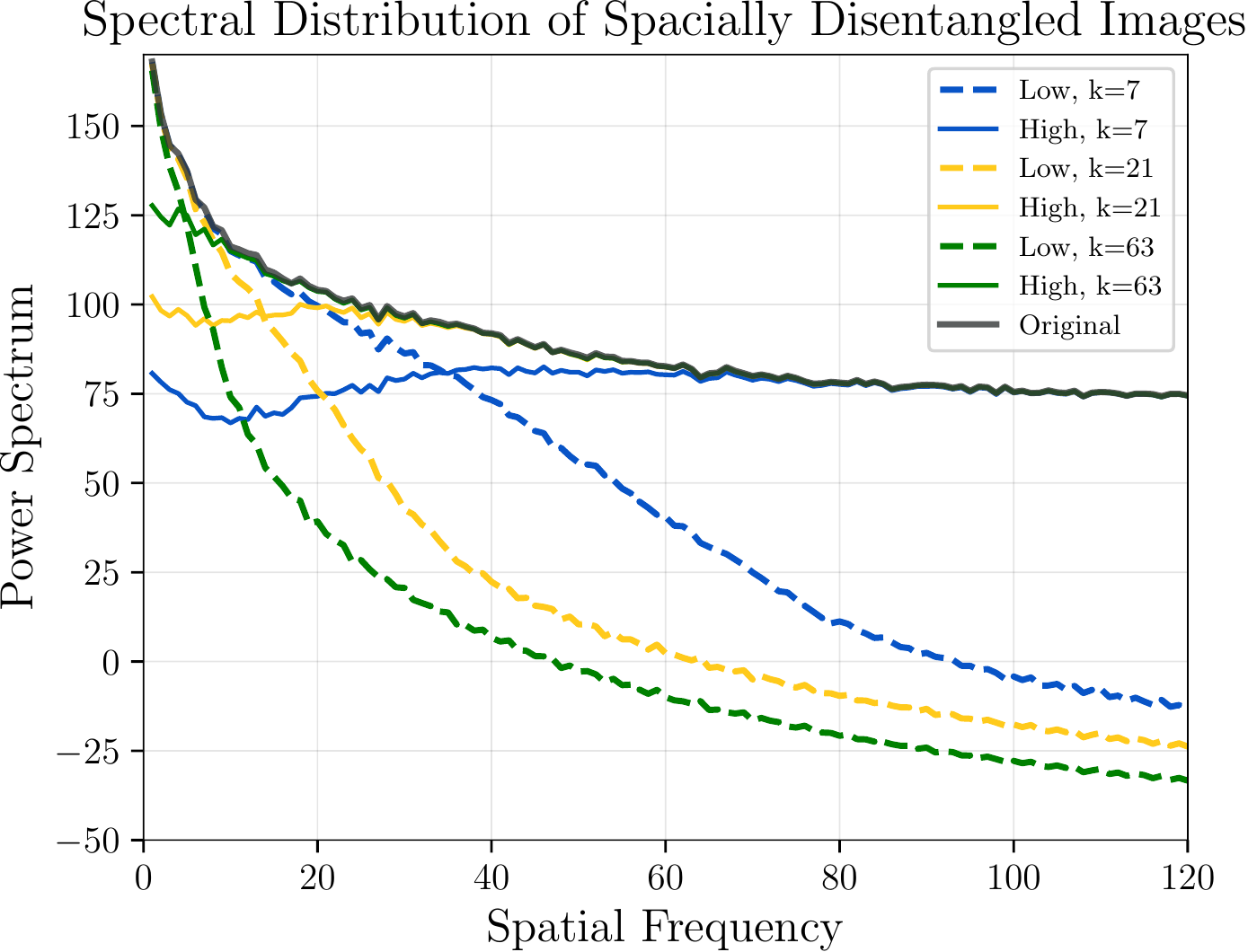}
	\end{center}
	\vspace{-2ex}
	\caption{\small Transforming the resulting high- and low-frequency images in Figure~\ref{fig:blur_compare} into the frequency power spectrum. The Gaussian kernel with kernel size $k=21$ could avoid the distortion in high-frequency and low-frequency regions. The power spectrum represents the energy distribution at each spatial frequency. }
	\vspace{-2ex}
	\label{fig:mask}
 \end{figure}

\paragraph{Gaussian kernel size} When  transforming the images in Figure~\ref{fig:blur_compare} into the spectrum space, the effects of the Gaussian kernel size could be clearly reflected in Figure~\ref{fig:mask}.
To be specific, a large kernel would cause severe distortion on the low-frequency band while a small kernel would not preserve much of the high-frequency information. In this work, we choose the kernel size $k=21$  for images with resolution 256$\times$256, which could appropriately separate the high/low-frequency information, demonstrated in both image space and spectral space distribution. Our experiments also show that FDIT is not sensitive to the selection of $k$ as long as it falls into a mild range.

\section{Experiments}
In this section, we evaluate our proposed method on two state-of-the-art image translation architectures, \ie, Swapping Autoencoder~\cite{park2020swapping}, StarGAN v2~\cite{choi2020stargan}, and one GAN inversion model, \ie, Image2StyleGAN~\cite{Abdal_2019_ICCV}. Extensive experimental results show that
\name{} not only better preserves the identity, but also enhances image quality. 

\vspace{-0.3cm}
\paragraph{Datasets.} 
We evaluate FDIT on the following five datasets:
(1) LSUN Church~\cite{yu2015lsun},
(2) CelebA-HQ~\cite{karras2017progressive},
(3) LSUN Bedroom~\cite{yu2015lsun},
(4) Flickr Mountains (100k self-collected images),
(5) Flickr Waterfalls (100k self-collected images). (6) Flickr Faces HQ (FFHQ) dataset~\cite{karras2019style}. All the images are trained and tested at $256 \times 256$ resolution except FFHQ, which is trained at $512 \times 512$, and finetuned at $1024 \times 1024$ resolution. For evaluation, we use a validation set that is separate from the training data. 

\subsection{Autoencoder}

Autoencoder is widely used as the backbone of the deep image translation task~\cite{Abdal_2019_ICCV,huang2018munit}.  We use state-of-the-art Swapping Autoencoder (SwapAE)~\cite{park2020swapping}, which is built on the backbone of StyleGAN2~\cite{Karras2019stylegan2}. Swap AE also uses the technique in PatchGAN~\cite{isola2017image} to further improve the texture transferring performance. We incorporate our proposed \name{} training objectives into the vanilla SwapAE.

\subsubsection{Reference-guided Image Synthesis}
\label{Reference-guided}

\begin{figure*}[t]
	\begin{center}
	  \includegraphics[width=0.92\textwidth]{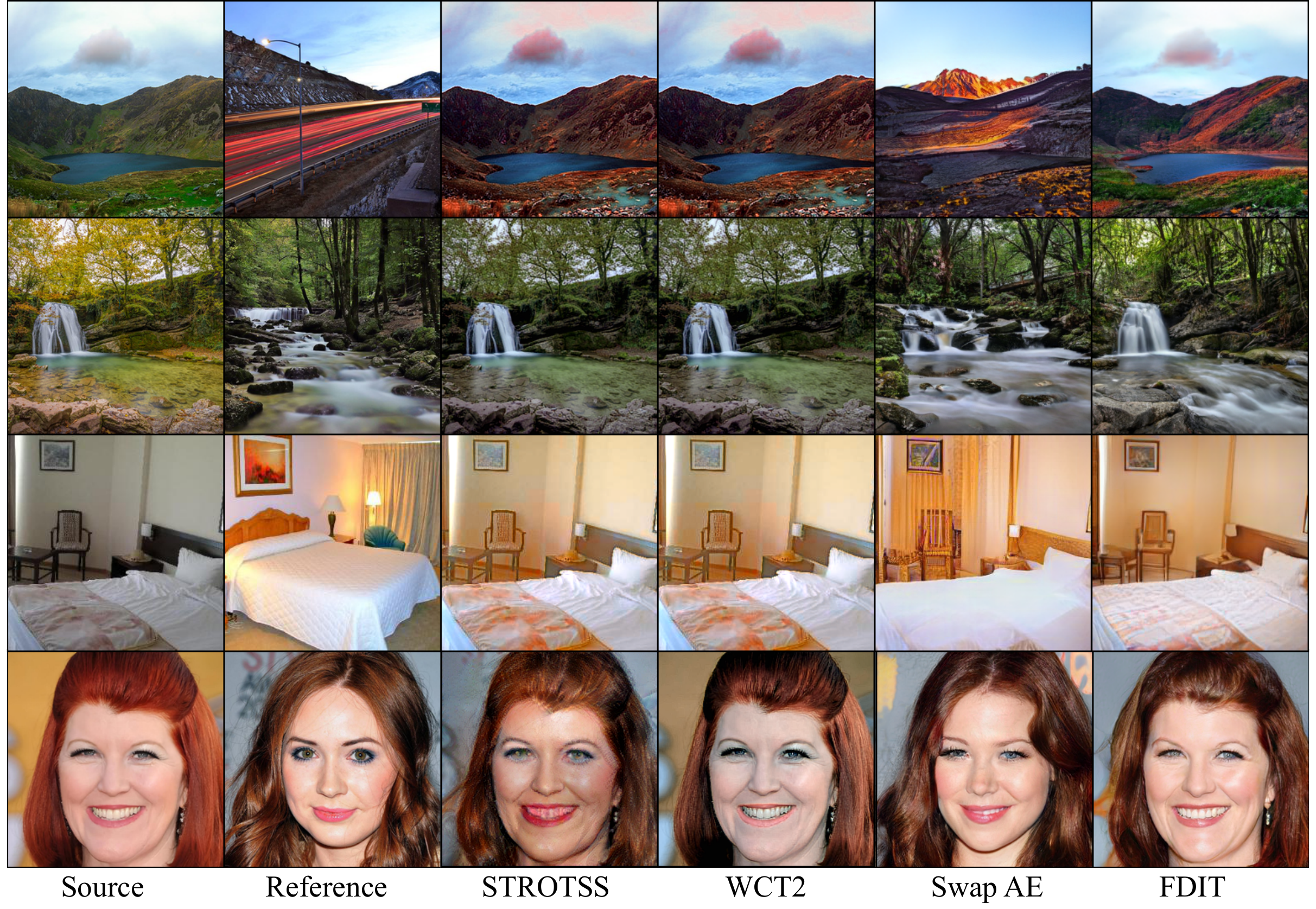}
	\end{center}
	\vspace{-5ex}
     \caption{\small Results across four diverse datasets, including Flicker Mountains, Flicker Waterfalls, LSUN Bedroom~\cite{yu2015lsun} , and CelebA-HQ~\cite{karras2017progressive}. Swap AE~\cite{park2020swapping} over-adapts to the reference image after image translation. In contrast, \textbf{FDIT} (ours) can better preserve  identity of the source image. Compared to STROTSS~\cite{Kolkin_2019_CVPR} and WCT2~\cite{Yoo_2019_ICCV}, FDIT can synthesize  photo-realistic images. Zoom in for details.}
	   \vspace{-2ex}
	\label{fig:face_trans}
 \end{figure*}

\paragraph{FDIT better preserves the identity with respect to the source image.}
We contrast the image translation performance using \name{} \vs vanilla SwapAE in  Figure~\ref{fig:church_trans} and Figure~\ref{fig:face_trans}. The vanilla SwapAE is unable to preserve the important identity of the source images, and over-adapts to the reference image. For example, the face identity is completely switched after translation, as seen in rows 4 of Figure~\ref{fig:face_trans}. SwapAE also fails to preserve the outline and the local sharp edges in the source image. As shown in Figure~\ref{fig:church_trans}, the outlines of the mountains are severely distorted.  Besides, the overall image composition has a large shift from the original source image. In contrast, using our method \name{}, the identity and structure of the swapped hybrid images are highly preserved. As shown in Figure~\ref{fig:church_trans} and Figure~\ref{fig:face_trans}, the overall sketches and local fine details are well preserved while the coloring, illumination, and even the weather are well transferred from the reference image (top rows of Figure~\ref{fig:church_trans}).

Lastly, we compare \name{} with
the state-of-the-art image stylization method STROTSS~\cite{Kolkin_2019_CVPR} and WCT2~\cite{Yoo_2019_ICCV}. Image stylization is a strong baseline as it emphasizes on the strict adherence to the source image. However, as shown in Figure~\ref{fig:face_trans}, WCT2 leads to poor transferability in image generation tasks. Despite strong identity-preservation, STROTSS and WCT2 are less flexible, and generate images that highly resemble the source image.  
In contrast, FDIT can both preserve the identity of the source image as well as maintain a high transfer capability. This further demonstrates the superiority of \name{} in image translation. 
 

\vspace{-2ex}
\paragraph{FDIT enhances the image generation quality.}
We show in Table~\ref{tab:table1} that \name{} can substantially improve the image quality while preserving the image content. We adopt the Fréchet Inception Distance (FID)~\cite{NIPS2017_8a1d6947} as the measure of image quality. Small values indicate better image quality. Details about Im2StyleGAN~\cite{Abdal_2019_ICCV} and StyleGAN2~\cite{Abdal_2019_ICCV} are shown in the supplementary material. \name{} achieves the lowest FID across all datasets. On average, FDIT could reduce the FID score by \textbf{5.6}$\%$ compared to the current state-of-the-art method.


\begin{table}[h]
   \footnotesize
  \begin{center}
  \begin{tabular}{ l|c|c|c|c}
      \toprule
        \diagbox{Method}{Dataset}~&~Church~&~Waterfalls&~FFHQ~&~CelebA-HQ \\
      \hline
        Im2StyleGAN~\cite{Abdal_2019_ICCV}~&~219.50~&~267.25&~123.13~&~- \\
        StyleGAN2~\cite{Abdal_2019_ICCV}~&~57.54~&~ 57.46&~81.44~&~- \\
        Swap AE~\cite{park2020swapping}~&~52.34~&~ 50.90&~59.83~&~43.47 \\
        \hline
          \name{} (ours)~&~\textbf{48.21}~&~\textbf{48.76}&~\textbf{55.96}~&~\textbf{42.02} \\
       \bottomrule
  \end{tabular}
  \end{center}
  \vspace{-5ex}
  \caption{\small Comparison of FID score on four diverse datasets: LSUN Church,  Waterfalls, FFHQ and CelebA-HQ. 
  }
  \vspace{-5ex}
  \label{tab:table1}
\end{table}

\subsubsection{Image Attributes Editing}


\paragraph{FDIT enables continuous interpolation between different domains.}
\label{para::Interpolation}
We show that FDIT enables image attribute editing task, which creates a series of smoothly changing images between two sets of distinct images~\cite{park2020swapping, shen2020interfacegan}. Vector arithmetic is one commonly used way to achieve this~\cite{shen2020interfacegan}. For example, we can sample $n$ images from each of the two target domains, and then compute the average difference of the vectors between these two sets of images: 

\begin{equation}
\begin{aligned}
  \mathbf{\hat{z}} &= \frac{1}{n}\sum\limits_{i=0}^{n} \mathbf{{z}}^\mathbf{d1}_i - \frac{1}{n}\sum\limits_{j=0}^{n} \mathbf{{z}}^\mathbf{d2}_j,
\end{aligned}
\label{equ:vector}
\end{equation}
where $\mathbf{{z^{d1}}}, \mathbf{{z^{d2}}}$ denote the latent code from two domains.



We perform interpolation on the style code while keeping the content code unchanged.
The generated images can be formalized as $\mathbf{x}_{\text{gen}}=G(\mathbf{z}^\text{source}, \mathbf{z}^\text{ref} + \theta \cdot \mathbf{\hat{z}} )$, where $\theta$ is the interpolation parameter. We show results on CelebA-HQ dataset in Supplementary material. 
FDIT performs image editing towards the target domain while strictly adhering to the content of the source image. Compared to the vanilla Swapping Autoencoder and StarGAN v2, our results demonstrate the better disentanglement ability of unique image attributes and identity characteristics. We also verify the disentangled semantic latent vectors using Principal Component Analysis (PCA). The implementation details and the identity-preserving results are shown in the supplementary materials. 


\subsection{Ablation Study}
\label{sec:ablation}
\noindent\textbf{Pixel and Fourier space losses are complementary.} To better understand our method, we isolate the effect of \emph{pixel space loss} and \emph{Fourier spectral space loss}. The results on the LSUN Church dataset are summarized in Table~\ref{tab:table-ablation}. The vanilla SwapAE is equivalent to having neither loss terms, which yields the FID score of 52.34. Using pixel space frequency loss reduces the FID score to 49.47. Our method is most effective when combining both pixel-space and Fourier-space loss terms, achieving the FID score of 48.21. Our ablation signifies the importance of using frequency-based training objectives.

\begin{table}[hbpt]
    \begin{center}
    \begin{tabular}{ c c |c }
        \toprule
         \multicolumn{2}{c|} {Loss terms}&\\
         Pixel
         space& Fourier space& FID $\downarrow$  \\
         \hline
           \xmark&\xmark&52.34\\
            \checkmark&\xmark&49.47\\
            \xmark &\checkmark&49.62\\
            \checkmark&\checkmark&\textbf{48.21} \\
        \bottomrule
    \end{tabular}
    \end{center}
    \vspace{-4ex}
    \caption{\small Ablation study on the effect of pixel space loss and Fourier spectral space loss. Evaluations are based on the LSUN Church dataset.}
    \label{tab:table-ablation}
 \end{table}

\subsection{GAN Inversion}

\noindent\textbf{FDIT improves reconstruction quality in GAN inversion.} We evaluate the efficacy of \name{} on the GAN inversion task, which maps the real images into the noise latent vectors. In particular, Image2StyleGAN\cite{Abdal_2019_ICCV} serves as a strong baseline, which performs reconstruction between the real image and the generated images via iterative optimization over the latent vector.

\begin{figure}[h]
  \centering     
  \vspace{-0.3cm}
  \includegraphics[width=0.47\textwidth]{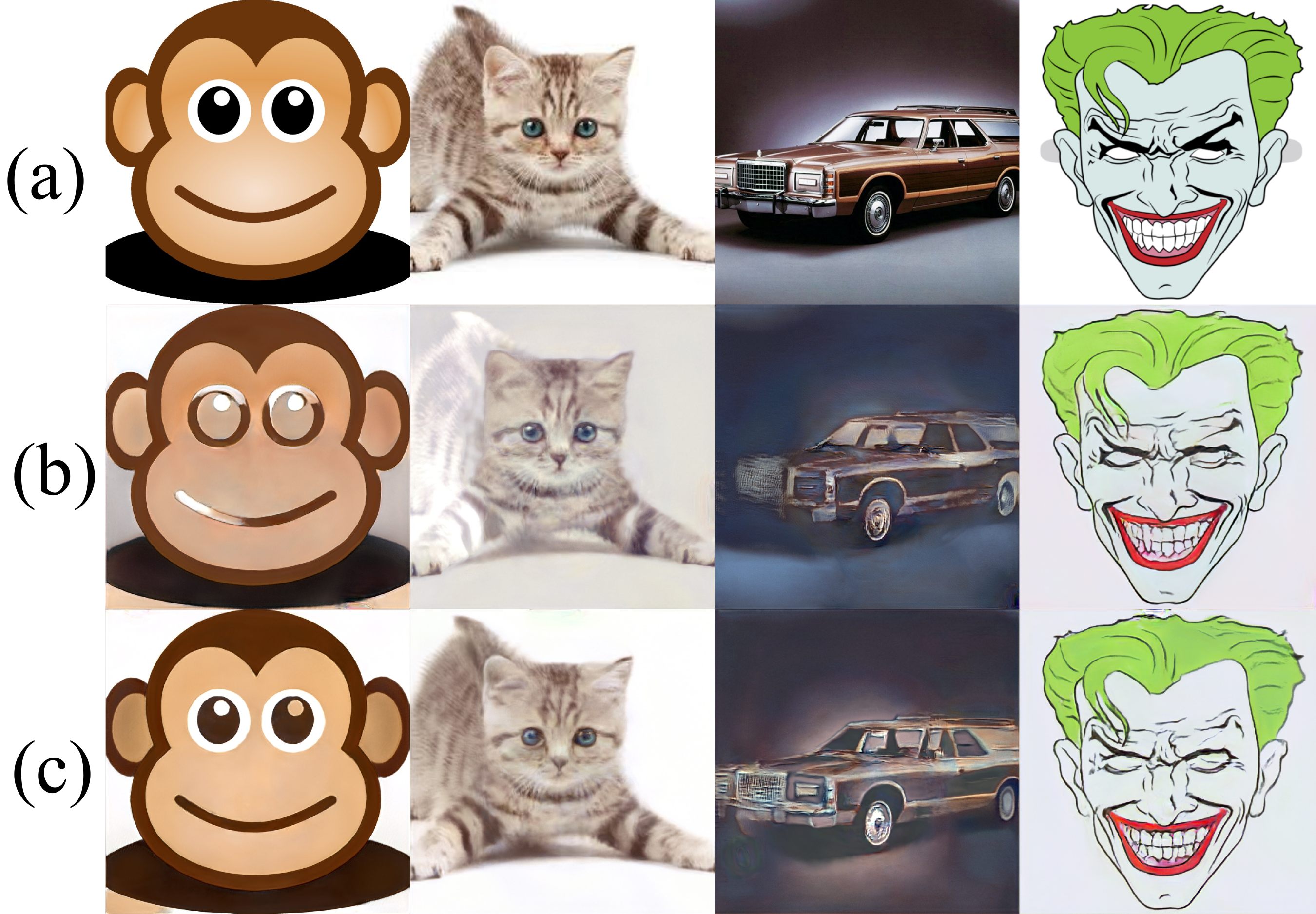}   
  \caption{\small GAN inversion results on high resolution images (1024$\times$1024). We compare  (a) high-resolution source images, (b) Image2StyleGAN\cite{Abdal_2019_ICCV} results and (c) inverted images by FDIT (ours). FDIT better maintains fine details and visual quality. } 
  \label{fig:img2stylegan}  %
  \vspace{-2ex}
\end{figure}

We adopt the same architecture, however impose our frequency-based reconstruction loss. The inversion results are  shown in Figure~\ref{fig:img2stylegan}. On high-resolution ($1024\times 1024$) images, the quality of the inverted images is improved across all scenes. \name{} better preserves the overall structure, fine details, and color distribution. We further measure the performance quantitatively,  summarizing the results in Table~\ref{tab:table_inversion}. Under different metrics (MSE, MAE, PSNR, SSIM), our method \name{} outperforms Image2StyleGAN. 

\begin{table}[bpt]
  \begin{center}
  \small
  \begin{tabular}{ l|c|c}
      \toprule
        \diagbox{Metrics}{Method}~&~Image2StyleGAN~&~FDIT \\
      \hline
        MSE $\downarrow$~&~0.0226~&~\textbf{0.0205} \\
        MAE $\downarrow$~&~0.0969~&~\textbf{0.0860} \\
        PSNR $\uparrow$~&~19.626~&~\textbf{20.466} \\
        SSIM $\uparrow$~&~0.6160~&~\textbf{0.6218} \\
       \bottomrule
  \end{tabular}
  \end{center}
  \vspace{-4ex}
  \caption{\small GAN inversion performance comparison, measured by the image reconstruction quality  between Image2StyleGAN and FDIT (ours). Evaluation metrics includes mean-square error~(MSE), mean absolute error~(MAE), peak signal-to-noise ratio~(PSNR)~\cite{de2003improved}, and SSIM~\cite{wang2004image}. $\uparrow$ means that higher value represents better image quality, and vice versa.  
  }
  \vspace{-4ex}
  \label{tab:table_inversion}
\end{table}




\subsection{StarGAN v2}

StarGAN v2 is another state-of-the-art image translation model which can generate image hybrids guided by either reference images or latent noises. 
Similar to the autoencoder-based network, we can optimize the StarGAN v2 framework with our frequency-based losses.
In order to validate FDIT in a stricter condition, we construct a CelebA-HQ-Smile dataset based on the smiling attribute from CelebA-HQ dataset. The style refers to whether that person smiles,  and  the content refers to the identity.

\begin{figure}[bpt]
  \centering      
  \includegraphics[width=0.45\textwidth]{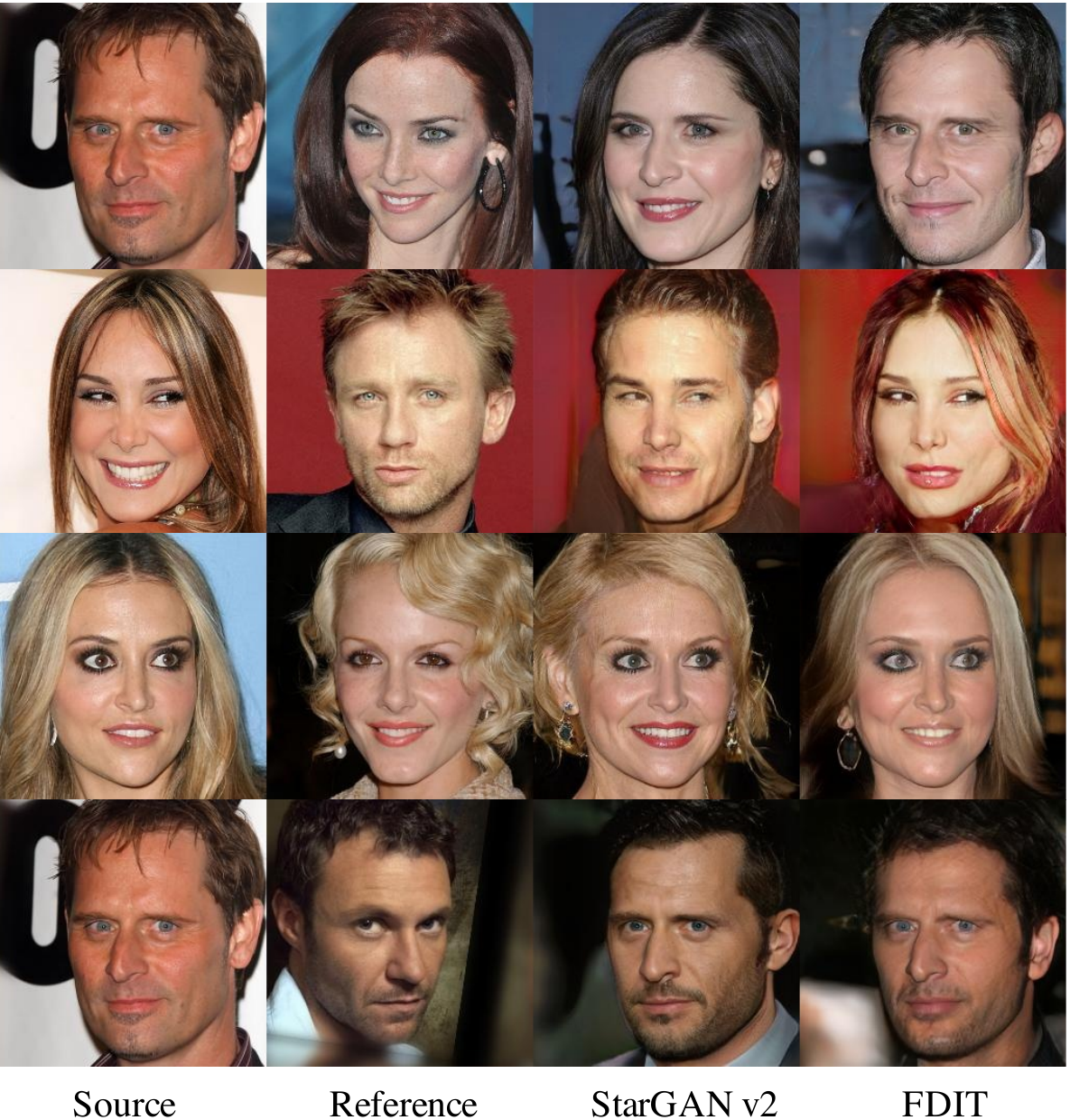}  
  \caption{\small Compared to vanilla StarGAN v2~\cite{choi2020stargan}, FDIT achieves much better identity-preserving ability.}
  \vspace{-0.3cm}
  \label{fig:stargan}  
\end{figure}

Several salient observations can be drawn from Figure~\ref{fig:stargan}. First, FDIT can highly preserve the gender identity; whereas the vanilla StarGAN v2 model would change the resulting gender according to the reference image (\eg first and second row). Secondly, the image quality of FDIT is better, where FID is improved from  {17.32} to {16.86}. Thirdly, our model can change the smiling attribute while maintaining other facial features strictly. For example, as shown in the third row, StarGAN v2 undesirably changes the hairstyle from straight (source) to curly (reference), whereas \name{} maintains the same hairstyle.

\subsection{User Study}
We conduct a user study to qualitatively measure the generated images. Specifically, we employ the two-alternative forced-choice setting, which was commonly used to train Learned Perceptual Image Patch Similarity~(LPIPS)~\cite{zhang2018perceptual} and to evaluate style transfer methods.
We provide users with the source image, reference image, images generated by FDIT, and the baseline SOTA models. Each user is forced to choose which of the two  image hybrids 1) better preserves the identity characteristics, and 2) has better image quality. We collected a total of 2,058 user preferences across 5 diverse datasets. Results are summarized in Table~\ref{tab:table_user}. On average, \textbf{75.40\%} of preferences are given to FDIT for identity preserving; and \textbf{64.39\%} of answers indicate FDIT produces more photo-realistic images. 

Furthermore, comparing to StarGAN v2, {57.14\%} user preferences are given to FDIT for better content preservation; {53.34\%} user preferences indicate that  FDIT produces better image quality compared to Image2StyleGAN. Therefore, the user study also verifies that {FDIT produces better identity-preserving and photo-realistic images.}

 \begin{table}[hbpt]
    \vspace{-0.2cm}
    \begin{center}
    \small
    \begin{tabular}{ l|p{1.6cm}<{\centering}|c }
      \toprule
         \diagbox{Dataset}{Ratio(\%)}{Metric}~&~{\centering Identity Preserving }~&~Image Realism  \\
        \hline
        LSUN Church~&~63.27~&~57.14 \\
        LSUN Bedroom~&~71.43~&~ 78.57\\
        Flicker Mountains~&~80.10~&~ 66.84 \\
        Flicker Waterfalls~&~80.61~&~62.24\\
        CelebA-HQ~&~57.14~&~ 53.06\\
         \hline
            Average~&~\textbf{75.40}~&~\textbf{64.39} \\
        \bottomrule
    \end{tabular}
    \end{center}
    \vspace{-4ex}
    \caption{\small Results of the user study on five datasets, which shows the preference of FDIT over Swapping Autoencoder~\cite{park2020swapping} w.r.t identity preserving and image quality.}
    \vspace{-4ex}
    \label{tab:table_user}
 \end{table}


\section{Related work}

\paragraph{Generative adversarial networks (GAN).} GAN~\cite{goodfellow2014gan, NIPS2017_892c3b1c, pmlr-v70-arjovsky17a, brock2018large, 8237891, NIPS2016_8a3363ab}  has revolutionized revolutionized many computer vision tasks, such as super resolution~\cite{ledig2017srgan,wang2018esrgan}, colorization~\cite{kim2019tag2pix,yoo2019coloring}, and image synthesis~\cite{brock2018biggan,lucic2019s3gan,donahue2019bigbigan}. Early work~\cite{radford2015unsupervised, Huang_2017_CVPR} directly used the Gaussian noises as inputs to the generator. However, such an approach has unsatisfactory performance in generating photo-realistic images. Recent works significantly improved the image reality by injecting the noises hierarchically~\cite{karras2019style, Karras2019stylegan2} in the generator. These works adopt the adaptive instance normalization (AdaIN) module~\cite{huang2017arbitrary} for image stylization. 

\vspace{-0.3cm}
\paragraph{Image-to-image translation.} 
Image-to-image translation~\cite{zhu2017toward,  wang2018pix2pixHD} synthesizes images by following the style of a reference image while keeping the content of the source image. One way is to use the GAN inversion, which maps the input from the pixel space into the latent noises space via the optimization method~\cite{Abdal_2019_ICCV, 9157575, Karras2019stylegan2}. However, these methods are known to be computationally slow due to their iterative optimization process, which makes deployment in mobile devices difficult~\cite{Abdal_2019_ICCV}. Furthermore, the quality of the reconstructed images can be suboptimal.
Another approach is to utilize the conditional GAN (or autoencoder) to convert the input images into latent vectors~\cite{huang2018munit, choi2018stargan, choi2020stargan, park2020swapping,park2019SPADE, park2020contrastive}, making the image translation process much faster than GAN inversion. 
However, exiting state-of-the-art image translation models such as StarGAN v2~\cite{choi2020stargan} and Swapping Autoencoder~\cite{park2020swapping} can lose important structural characteristics of the source image. 
In this paper, we show that frequency-based information can effectively preserve the identity of the source image and enhance photo-realism. 


\vspace{-0.3cm}
\paragraph{Frequency domain in deep learning.} Frequency domain analysis is widely used in traditional image processing~\cite{heideman1984gauss, cooley1987re, van1992computational, johnson2006modified, gentleman1966fast}. The key idea of frequency analysis is to map the pixels from the Euclidean space to a frequency space, based on the changing speed in the spatial domain. Several works tried to bridge the connection between deep learning and frequency analysis~\cite{xu2020learning, chen2019drop, Xu_2020, 10.1007/978-3-030-36708-4_22, NIPS2016_36366388, NEURIPS2018_a9a6653e}. Chen~\etal~\cite{chen2019drop} and Xu~\etal~\cite{xu2020learning} showed that by incorporating frequency transformation, the neural network could be more efficient and effective. Wang~\etal~\cite{wang2020high} found that the high-frequency components are useful in explaining the generalization of neural networks. Recently, Durall~\etal~\cite{durall2020watch} observed that the images generated by GANs are heavily distorted in high-frequency parts, and they introduced a spectral regularization term to the loss function to alleviate this problem. Czolbe~\etal~\cite{watsongan} proposed a frequency-based reconstruction loss for VAE using discrete Fourier Transformation (DFT). However, this approach does not incorporate pixel space frequency information, and relies on a separate dataset to get its free parameters. In fact, no prior work has explored using frequency-domain analysis for the image-to-image translation task. In this work, we explicitly devise a novel frequency domain \emph{image translation} framework and demonstrate its superiority in performance.

\vspace{-0.3cm}
\paragraph{Neural style transfer.} Neural style transfer aims at transferring the low-level styles while strictly maintaining the content in the source image~\cite{Yoo_2019_ICCV, Kolkin_2019_CVPR, xhuang8237429,  WCT-NIPS-2017, li2018learning,  Li_2018_ECCV}. Typically, the texture is represented by the global image statistics while the content is controlled by the perception metric~\cite{Yoo_2019_ICCV, jing2019neural, zhang2018separating}. 
However, existing methods could only handle the local color transformation, making it hard to transform the overall style and semantics. More specifically, they struggle in the cross-domain image  translations, for example, gender transformation~\cite{Yoo_2019_ICCV}. In other words, despite strong identity-preservation ability, such methods are less flexible for the cross-domain translation and can generate images that highly resemble the source domain. In contrast, FDIT can both preserve the identity of the source images while maintaining a high domain transfer capability.

\section{Conclusion}

In this paper, we propose \emph{\namefull{}~({\name{}})}, a novel image translation framework that preserves the frequency information in both  {pixel space} and {Fourier spectral space}. Unlike the existing image translation models, FDIT directly uses high-frequency components to capture  object structure akin to the identity. 
Experimental results on five large-scale datasets and multiple tasks show that \name{} effectively preserves the identity of the source image while producing photo-realistic image hybrids. Extensive user study and ablations further validate the effectiveness of our approach both qualitatively and quantitatively.
We hope future research will increase the attention towards frequency-based approaches for image translation tasks.

\section{Acknowledgment}
Mu Cai and Yixuan Li are supported by funding from the Wisconsin Alumni Research Foundation (WARF). Gao Huang is supported in part by the National Key R\&D Program of China under Grant 2020AAA0105200,  the National Natural Science Foundation of China under Grants 62022048 and 61906106, the Institute for Guo Qiang of Tsinghua University and Beijing Academy of Artificial Intelligence.

{\small
\bibliographystyle{ieee_fullname}
\bibliography{citation}
}

\clearpage





\appendix
\onecolumn
\begin{center}
      {\large \bf {Frequency Domain Image Translation: More Photo-realistic, Better Identity-preserving
      
      (Supplementary Material)} \par}
     
      \vskip .5em
      \vspace*{12pt}
\end{center}

\section{ Image Attributes Editing Results }

We demonstrate the identity preserving capability and photo realism of \name{} under the image attribute editing task via continuous interpolation and unsupervised semantic vector discovery.

\subsection{Continuous interpolation  between Different Domains}

\begin{figure*}[h]
  \centering
   \begin{minipage}[c] {\linewidth} 
  \includegraphics[width=\linewidth]{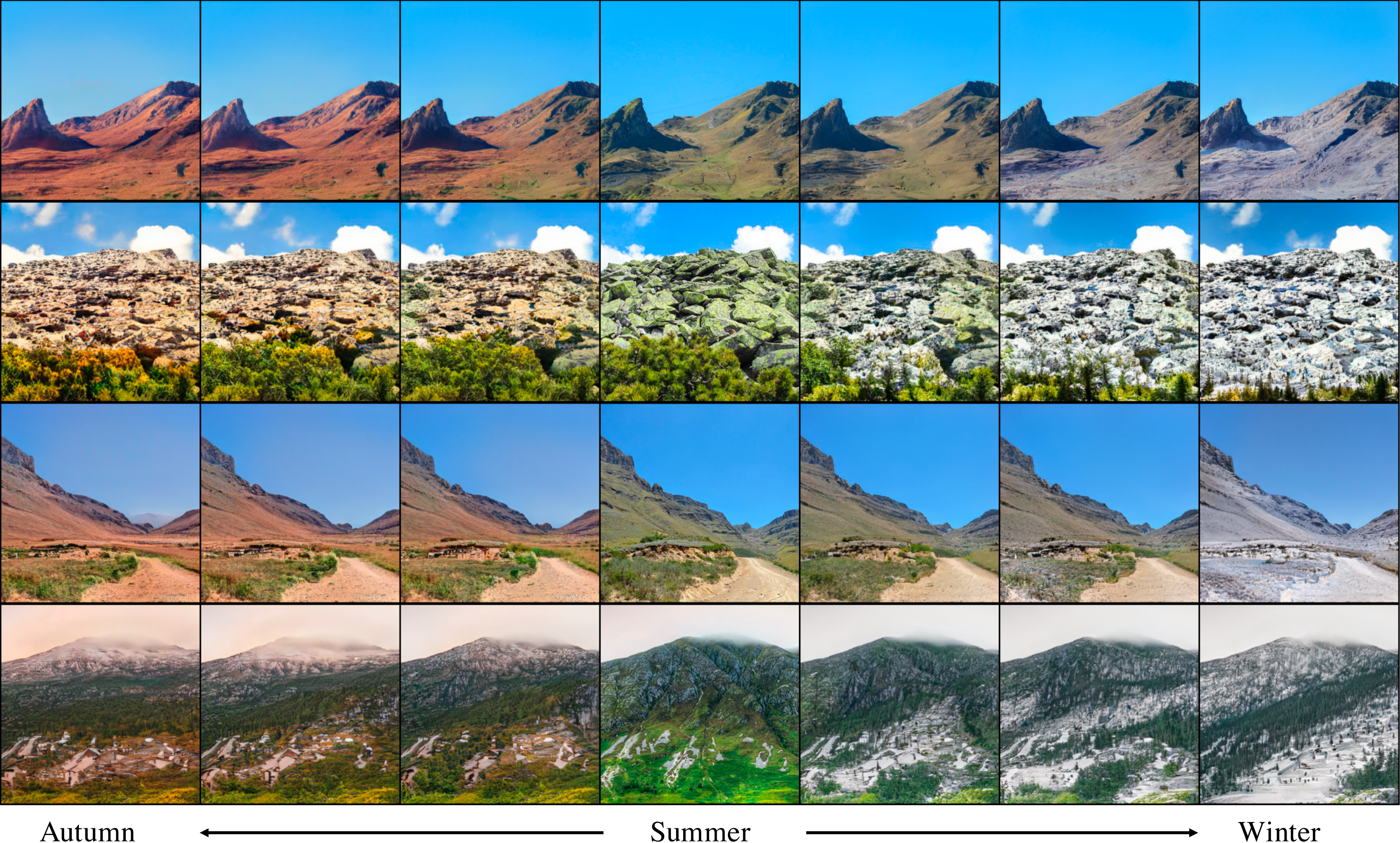}\\
   \end{minipage}
  \caption{Image attributes editing results of the LSUN mountain dataset~\cite{yu2015lsun} under the continuous interpolation. The central column denotes the source summer images, while the remaining columns denote the continuous interpolation images targeting at autumn and winter.} 
  \label{fig:church_winter} 
  \vspace{-2ex}
\end{figure*}

We show that FDIT can generate a series of smoothly changing images between two sets of distinct images. We perform interpolation on the style code while keeping the content code unchanged. Figure~\ref{fig:church_winter}  shows season transformation results using the Flicker Mountains dataset.
Our identity-preserving image hybrids demonstrate that
FDIT could achieve high-quality image editing performance towards the target domain while strictly adhering to the identity of the source image. 


\subsection{Unsupervised Semantic Vector Discovery for Image Editing}

Another way to conduct image editing is to discover the underlying semantics $\mathbf{\hat{z}}$  via an unsupervised way. Here we adopt the Principal Component Analysis~(PCA)~\cite{h2020ganspace} to achieve this goal, which could find the orthonormal components in the latent space. Similar to the continuous interpolation approach in our paper, when manipulating the style code using PCA,  a good image translation model would keep the content of the images as untouched as possible.
 
\begin{figure*}[h]
    \centering
    \includegraphics[width=1\linewidth]{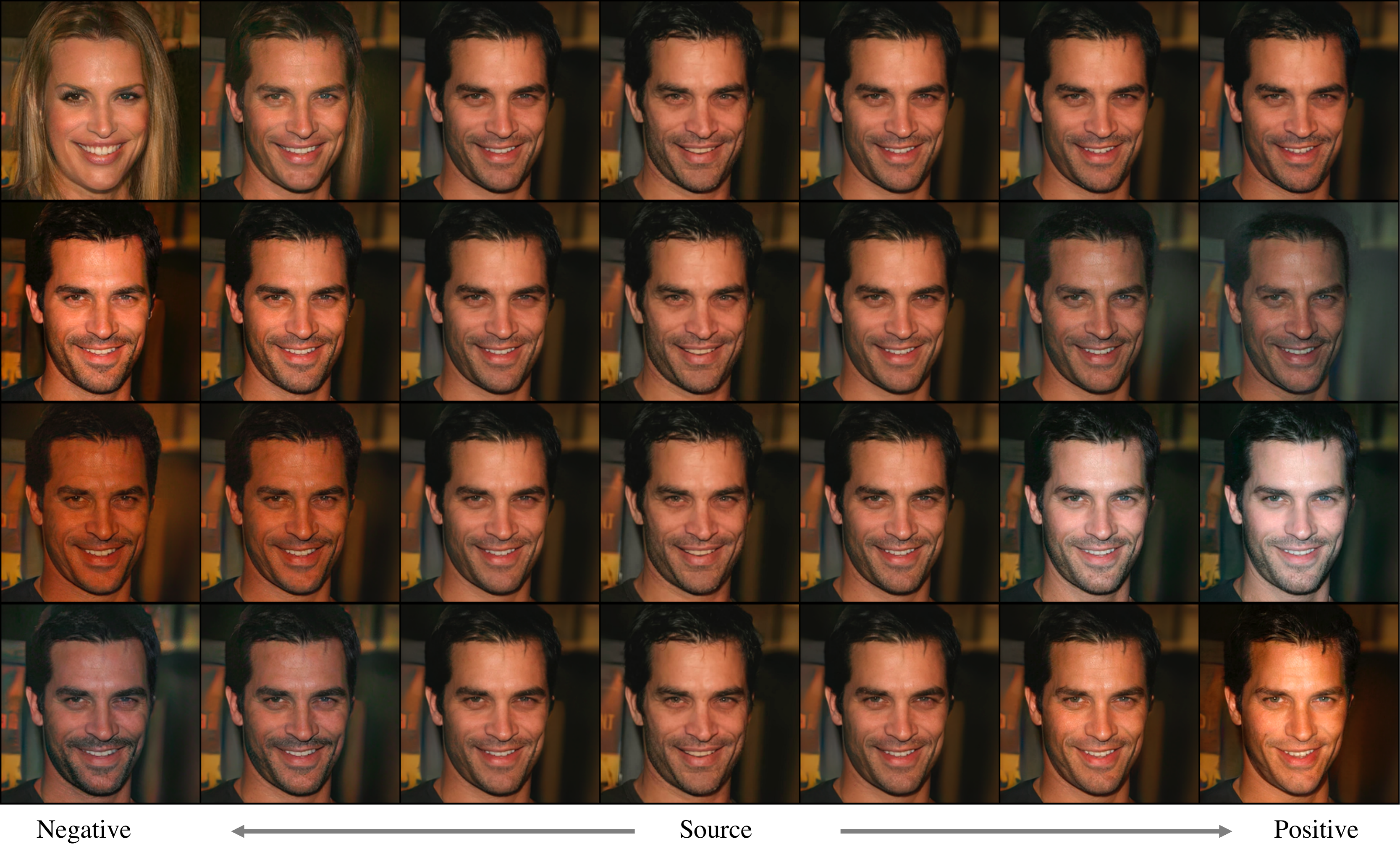}\\
      \caption{PCA-based image attributes editing results under the CelebA-HQ dataset. The central column denotes the source images, while within the remaining columns denote the  interpolation results of the orthonormal components along two directions.}
    \label{fig:pca}  %
    \end{figure*}

As shown in Fig.~\ref{fig:pca}, FDIT is once again demonstrated to be an identity-preserving model. Specifically, the identities are well maintained, while the only facial attributes such as illumination and hair color are changed.

\begin{figure}[b]
	\begin{center}
	\vspace{-0.6cm}
	  \includegraphics[width=\linewidth]{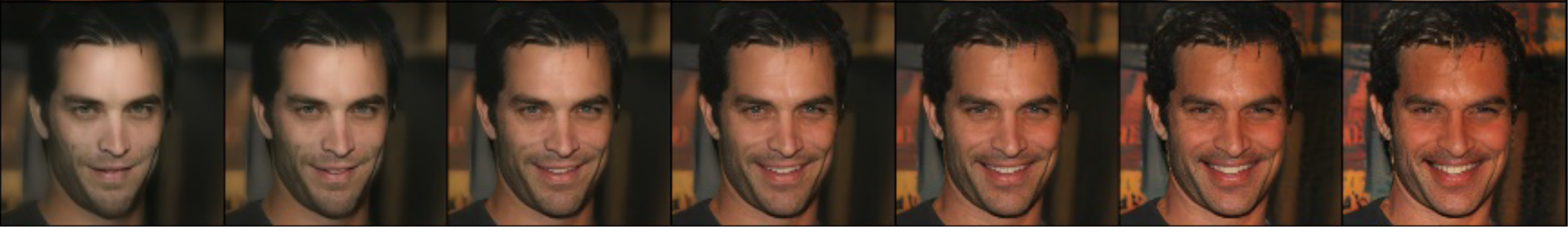} 
	\end{center}
	\vspace{-4.0ex} 
	   \caption{Image editing results using PCA on the full latent space. }
	   \vspace{-2ex}
	\label{fig:ablation_full_sapce}
 \end{figure}

We additionally show results of image editing in the \emph{full latent space} in Figure~\ref{fig:ablation_full_sapce}, which displays more variation. 

\section{Frequency Domain Image Translation Results}

We show the image generation results of the autoencoder based FDIT framework on LSUN Church~\cite{yu2015lsun}, CelebA-HQ~\cite{karras2017progressive}, Flickr Waterfalls, and LSUN Bedroom~\cite{yu2015lsun} in Figure~\ref{fig:all_img}.  \name{} framework achieves better performance in preserving the shape, which can be observed in the outline of the churches, the layout of the bedrooms, and the scene of the waterfalls.

\newcommand{\figsize}{0.485}

\begin{figure}[htbp]%
\centering
\begin{minipage}{\figsize \linewidth}
\centering
\subfigure[LSUN Church]{\includegraphics[width=\linewidth]{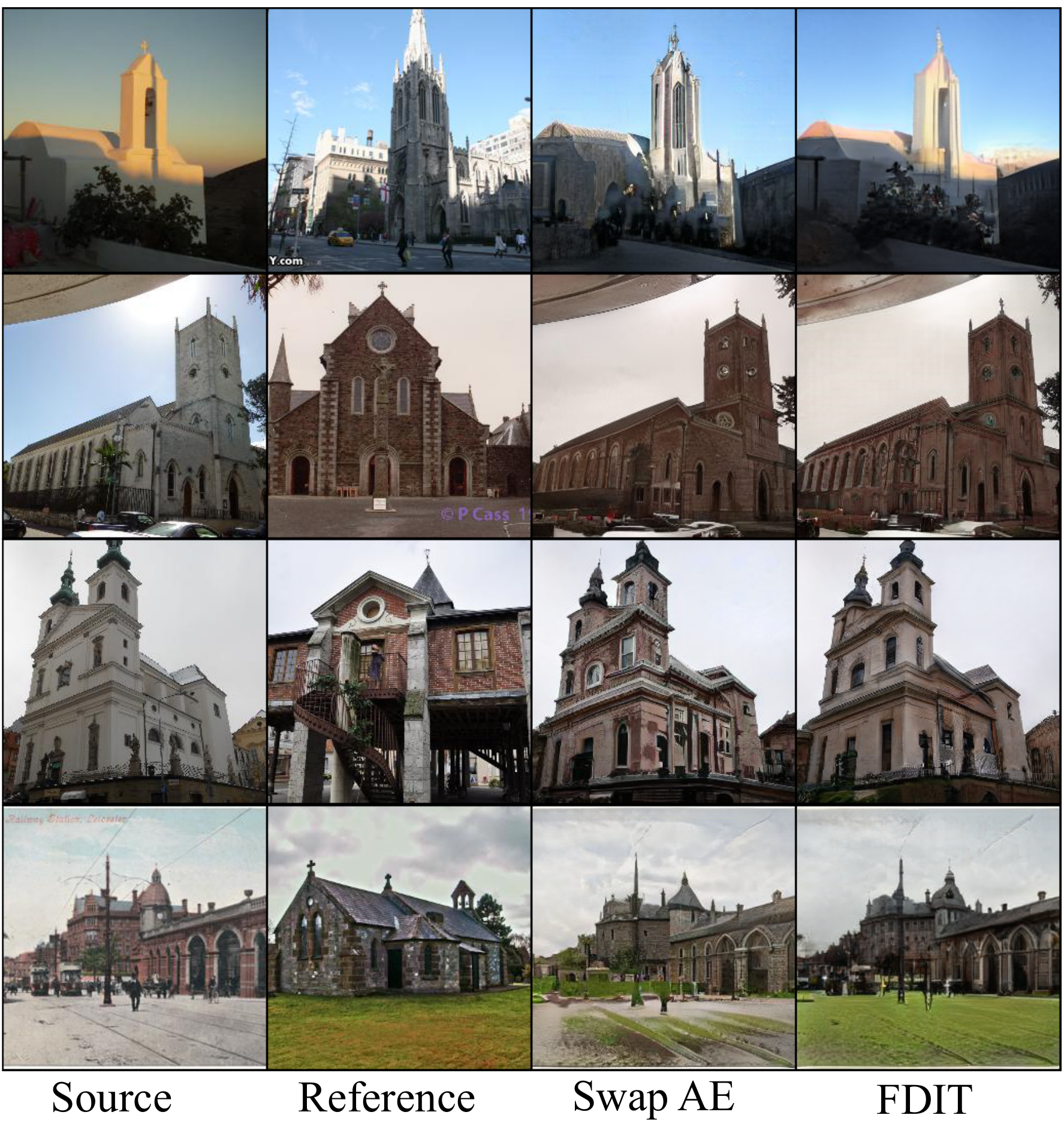}}
\end{minipage}\quad
\begin{minipage}{\figsize \linewidth}
\centering
\subfigure[CelebA-HQ]{\includegraphics[width=\linewidth]{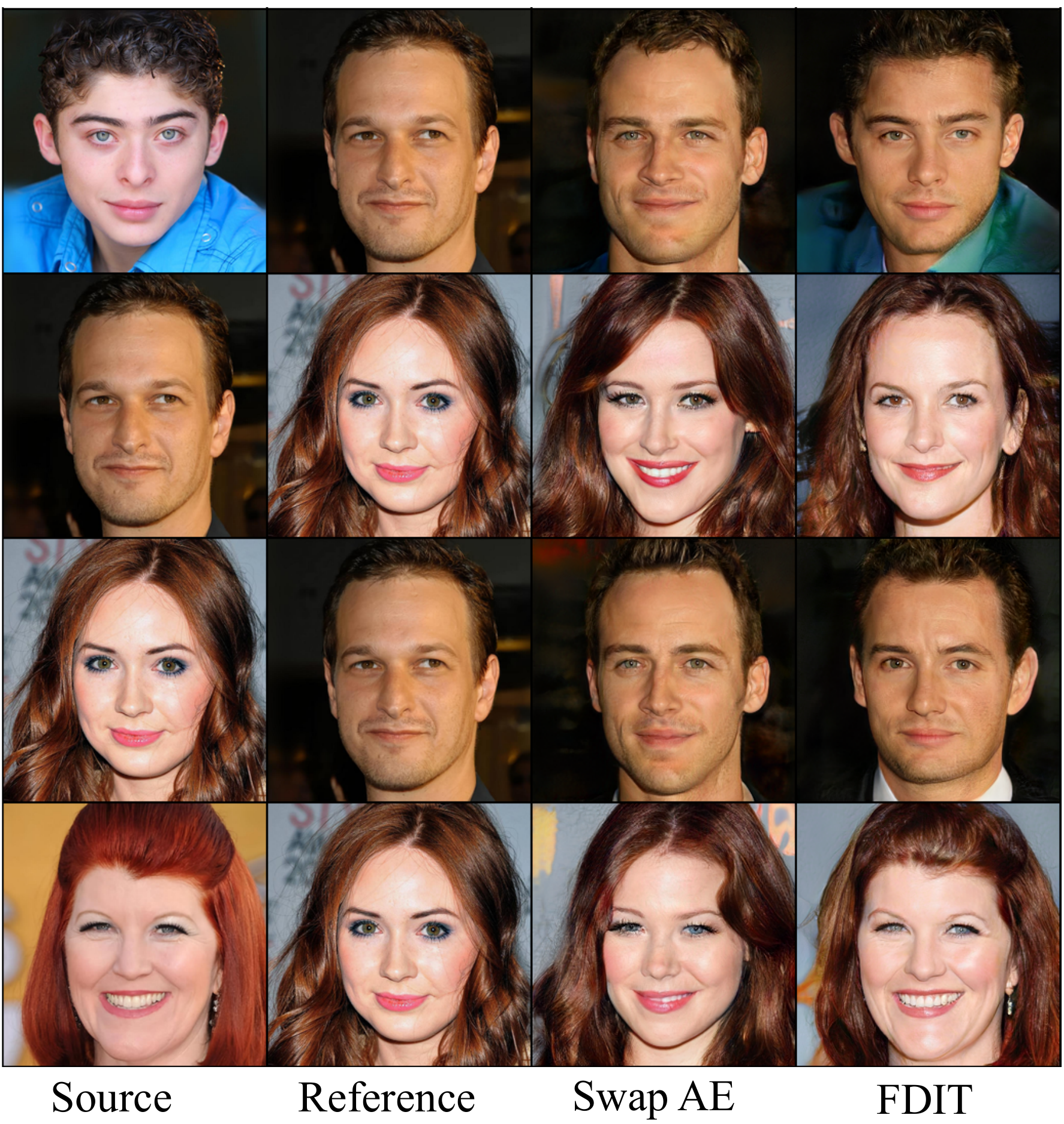}}
\end{minipage}

\begin{minipage}{\figsize \linewidth}
\centering
\subfigure[Flicker Waterfalls]{\includegraphics[width=\linewidth]{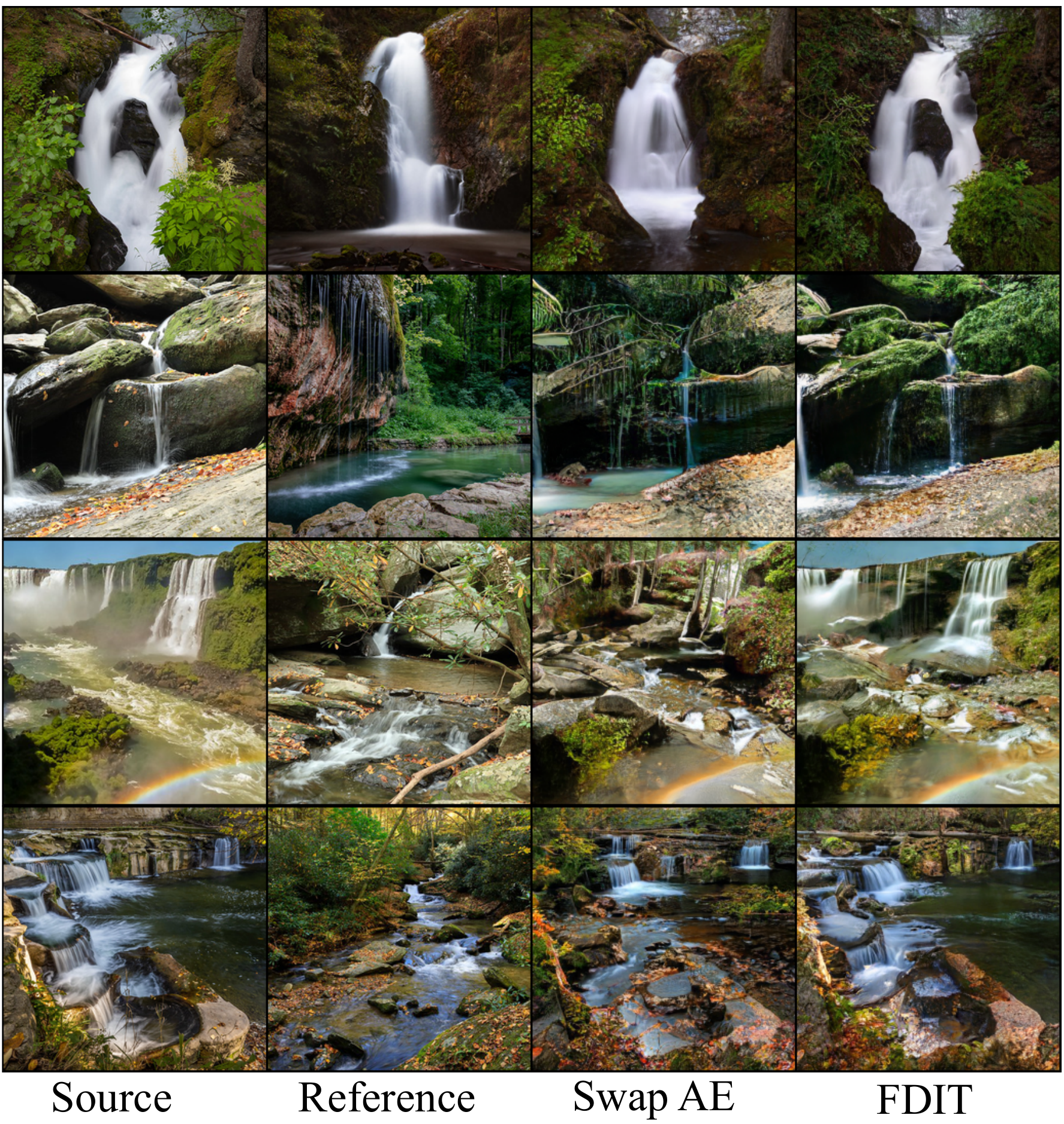}}
\end{minipage}\quad
\begin{minipage}{\figsize \linewidth}
\centering
\subfigure[LSUN Bedroom]{\includegraphics[width=\linewidth]{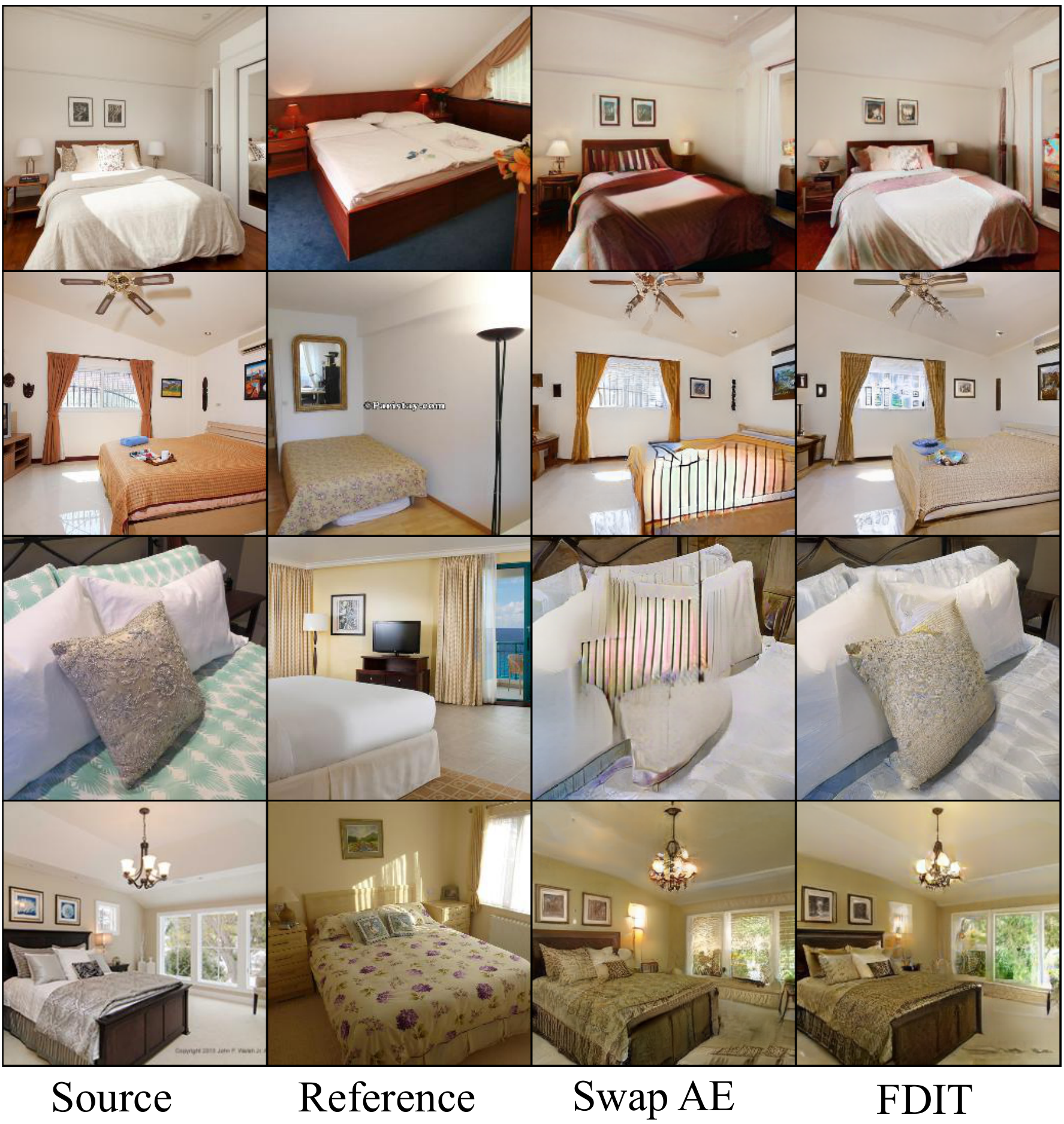}}
\end{minipage}
\caption{Image translation results under the (a) LSUN Church, (b) CelebA-HQ, (c) Flicker Waterfalls, and (d) LSUN Bedroom dataset. Four columns denote the source images, reference images, and the generated images of Swapping Autoencoder~\cite{park2020swapping} and FDIT, respectively.}
\label{fig:all_img}  
\end{figure}

\section{ Constructing the Flicker Dataset}

We collect the large-scale Flicker Mountains dataset and Flicker Waterfalls dataset from \url{flickr.com}. Each dataset contains 100,000 training images. 

\section{Training Details}

Our \namefull~(\name) framework is composed of the  \emph{pixel space} and  \emph{Fourier frequency space} losses, which can be conveniently implemented for existing image translation models. For fair comparison, we keep all training and evaluation settings the same as the baselines (Swapping Autoencoder\footnote{\url{https://github.com/rosinality/swapping-autoencoder-pytorch}}~\cite{park2020swapping}, StarGAN v2\footnote{\url{https://github.com/clovaai/stargan-v2}}~\cite{choi2020stargan}, and Image2StyleGAN\footnote{\url{https://github.com/pacifinapacific/StyleGAN_LatentEditor}}~\cite{Abdal_2019_ICCV}). All experiments are conducted on the Tesla V100 GPU.

 \paragraph{Swapping Autoencoder~\cite{park2020swapping}.} The encoder-decoder backbone is built on StyleGAN2~\cite{Karras2019stylegan2}. We train the model on the 32GB Tesla V100 GPU, where the batch size is 16 for images of 256$\times$256 resolution, and 4 for images of $1024\times1024$ resolution.  During training, a batch of $n$ images are fed into the model, where $\frac{n}{2}$ reconstructed images and $\frac{n}{2}$ image hybrids would be produced. We adopt Adam~\cite{adam} optimizer where $\beta_1 = 0, \beta_2=0.99$. The learning rate is set to be 0.002. The reconstructed quality is supervised by $L_1$ loss. The discriminator is optimized using the adversarial loss~\cite{goodfellow2014gan}. A patch discriminator is utilized to enhance the texture transferring ability \textit{w.r.t.} reference images.
 
 \paragraph{StarGAN v2~\cite{choi2020stargan}.} We use the official implementation in StarGAN v2, where the backbone is built with ResBlocks~\cite{he2016deep}. The batch size is set to be 8. Adam~\cite{adam} optimizer is adopted where $\beta_1 = 0, \beta_2=0.99$.  The learning rate for the encoder, generator, and discriminator is set to be $10^{-4}$. In the evaluation stage, we utilize the exponential moving averages over encoder and generator. 
 
\paragraph{Image2StyleGAN v2~\cite{Abdal_2019_ICCV}.}  We adopt the Adam optimizer with the learning rate of 0.01, $\beta_1 = 0.9, \beta_2=0.999$, and $\epsilon=1 e^{-8}$ in the experiments. We use 5000 gradient descent steps to obtain the GAN-inversion images.

\section{Details of Image2StyleGAN and StyleGAN2 results in Table 1.} 

Both Im2StyleGAN~\cite{Abdal_2019_ICCV} and StyleGAN2~\cite{Abdal_2019_ICCV} invert the image from the training domain, then use the mixed latent representations to create image hybrids. Image2StyleGAN  adopts the iterative optimization on the '$ W^{+}$-space'  to project images using the StyleGAN-v1 backbone; while StyleGAN2 utilizes an LPIPS-based projector under the StyleGAN-v2 backbone. 

\section{The qualitative results for Section 4.2}

\begin{figure}[t]
	\begin{center}
	  \includegraphics[width=0.8\textwidth]{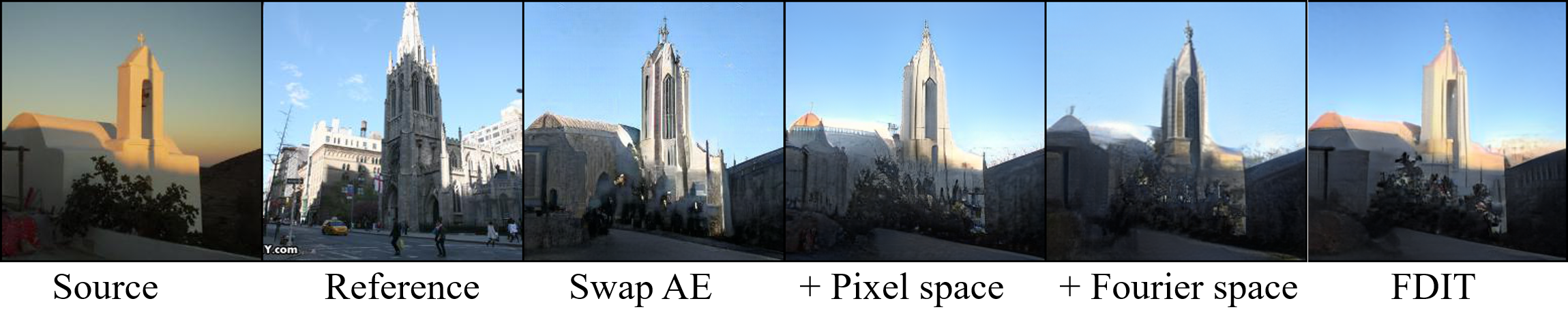}
	\end{center}
	\vspace{-4ex}
	   \caption{Image translation results of the Flicker mountains dataset. From left column to right: we show the source images, reference images, the generated images using Swap AE, with pixel space loss, with Fourier space loss,  and with both (FDIT), respectively. }
	   \vspace{-2ex}
	\label{fig:ablation_visual}
 \end{figure}

The qualitative results are shown in Figure~\ref{fig:ablation_visual}, where FDIT shows better identity preservation than using only pixel or Fourier loss. For example, using only Fourier loss preserves the identity but loses some style consistency in the pixel space.

\end{document}